\documentclass[journal]{IEEEtran}
\usepackage{graphicx}
\usepackage{lipsum}
\usepackage{tikz}
\usepackage{hyperref}

\ifCLASSINFOpdf
\else
\fi
\begin{document}
%
\title{Robustness Testing of Multi-Modal Models in \\ Varied Home Environments for Assistive Robots}
%
%
%

\author{Lea~Hirlimann,
        Shengqiang~Zhang,
        ~Hinrich~Schütze and
        Philipp~Wicke
\thanks{Corresponding author e-mail: philipp.wicke@lmu.de}
\thanks{All authors are with the Institute for Information and Language Processing at Ludwig Maximilian University of Munich. P. Wicke, S. Zhang and H. Schütze are affiliated with the Munich Center of Machine Learning (MCML)}
\thanks{Manuscript submitted June 19, 2024}}

%
%

\markboth{GERIATRONICS SUMMIT 2024, JULY 09 - 10, GARMISCH-PARTENKIRCHEN CONGRESS CENTER}%
{Shell \MakeLowercase{\textit{et al.}}: Bare Demo of IEEEtran.cls for IEEE Journals}
%



\maketitle

\begin{abstract}
The development of assistive robotic agents to support household tasks is advancing, yet the underlying models often operate in virtual settings that do not reflect real-world complexity. For assistive care robots to be effective in diverse environments, their models must be robust and integrate multiple modalities. Consider a caretaker needing assistance in a dimly lit room or navigating around a newly installed glass door. Models relying solely on visual input might fail in low light, while those using depth information could avoid the door. This demonstrates the necessity for models that can process various sensory inputs. Our ongoing study evaluates state-of-the-art robotic models in the AI2Thor virtual environment. We introduce \textit{disturbances}, such as dimmed lighting and mirrored walls, to assess their impact on modalities like movement or vision, and object recognition. Our goal is to gather input from the Geriatronics community to understand and model the challenges faced by practitioners.
\end{abstract}


%
\IEEEpeerreviewmaketitle

\section{Introduction}

Training robotic agents in virtual environments \cite{ye2022rcare,kolve2017ai2,james2020rlbench} is a common practice before deploying them in real-world settings \cite{bucker2022reshaping,bucker2023latte}. This method is both safe and effective. It ensures safety by preventing any risk to individuals, especially the elderly, who might rely on these robots for assistance and may not respond swiftly to dangerous situations. It is also effective because virtual environments allow extensive testing through thousands of different iterations and scenarios, thoroughly evaluating the robot's performance \cite{zhang2023lohoravens, wicke2023towards}. However, simulations inevitably simplify real-world physics and problems, which can lead to challenges when these robotic systems are implemented in real-world assistance contexts.

Many modern robotic agents use large transformer models trained on vision and language (VLMs \cite{radford2021learning}) to perform household tasks \cite{brohan2023can,driess2023palm}. These systems are often the result of end-to-end training, and in their deployment they function from task instruction to execution. While integrating multiple modalities — such as visual and auditory inputs — can enhance robot performance, it remains unclear which specific modalities contribute most to task success and how robust these modalities are when deployed in real-world scenarios. Researchers often use ablation studies to determine the impact of training with and without certain modalities \cite{padalkar2023open,shah2023lm}. However, there is a lack of direct testing on the robustness of these modalities under varying conditions.

This study aims to investigate the robustness of multi-modal models in varied home environments, focusing on assistive robots. We use the AI2Thor environment \cite{kolve2017ai2} to simulate \textit{disturbances} that affect semantic inputs in different ways. By creating a series of tasks for state-of-the-art robotic agents and introducing various \textit{disturbances}, we hope to identify potential issues that could arise in real-world deployment. This approach will also help us understand the dependencies of these models when different modalities are challenged. \\

Consider a scenario where a caretaker instructs a robot to fetch a bottle of wine from the kitchen, but one of the lightbulbs in the kitchen is broken, resulting in low light conditions. A robot that heavily relies on visual data might fail to complete the task. Conversely, if the robot has a semantic map of the kitchen and follows pre-learned paths, it might succeed without relying on vision, highlighting the importance of locomotion and path-following in this case. Further complicating the scenario, imagine a new mirror in the kitchen reflecting the image of a bottle. A robot equipped with depth perception might ignore the reflection, whereas one relying solely on object recognition could mistake the reflection for the actual bottle and grasp for the mirror.

By systematically disturbing semantic inputs in controlled virtual environments, we aim to identify potential weaknesses in current multi-modal models before they are deployed in real-world settings. Additionally, our findings could reveal intricate dependencies within these models when exposed to different disturbances. We hope this research will foster greater collaboration with practitioners in the field of geriatrics, enhancing our understanding of the challenges in deploying assistive robots and improving their reliability and safety in practical applications. Our key contributions of this preliminary work are summarised as follows: \\

\begin{itemize}
    \item Introducing the notion of \textit{disturbances} as a methodology which allows researchers to assess the robustness of multi-modal models in virtual environments. \\

    \item Contextualising robustness, disturbances and multi-modal models for Geriatronics research by means of a literature review and a comparative overview. \\

    \item An outline of planned research including suggested disturbances, tasks, environments and models. \\
\end{itemize}

\section{Related Works}

\paragraph{Simulation Environments}
The study of robotic agents in virtual environments has been significantly advanced by benchmarks like ALFRED (Action Learning From Realistic Environments and Directives). ALFRED, introduced by \cite{shridhar2020alfred}, leverages egocentric vision within the AI2Thor environment \cite{kolve2017ai2}, combined with natural language instructions to guide household tasks. The tasks are divided into high-level descriptions and low-level sequences, requiring navigation, object detection, interaction (e.g. picking objects up), and state changes (e.g. heating objects). The dataset includes seven distinct task types, ranging from pick and place to examining objects in light. For each task, detailed goal and instruction annotations were provided, resulting in over 8k expert demonstrations averaging 50 steps each. Ablation studies showed that the integration of vision and language modalities significantly improved task performance, highlighting the importance of multi-modal inputs in achieving higher success rates and robustness in diverse environments.

The AI2Thor simulation framework, described by \cite{kolve2017ai2}, plays a crucial role in the development and testing of robotic agents. AI2Thor offers a versatile platform for robotic navigation and interaction, with scenes ranging from individual rooms to multi-story houses. The framework includes four primary scene collections: iThor, ProcThor, RoboThor, and ArchitectThor. These simulations run on a Unity back-end, providing agents with metadata on action success, camera images, and object positions. The iTHOR scenes, which consist of single rooms like kitchens and living areas, are extensively used in the ALFRED benchmark. Additionally, expanded scene collections such as RoboThor and ArchitectThor offer more complex environments, while ProcThor features procedurally generated large-scale scenes. These enhancements in the simulation platform contribute to more robust and scalable training environments for robotic agents. Due to the available Unity back-end, AI2Thor is our preferred choice of environment, because we can easily modify the virtual space with disturbances through the game engine, while connecting these modified versions with household tasks provided by AI2Thor. \\

\paragraph{Modalities for Model Training}
The application of large language models (LLMs) in robotics has shown promising advancements in decision-making and planning capabilities. As noted by \cite{chen2023towards}, LLMs enhance reasoning and planning through methods like Chain of Thought \cite{wei2022chain} and problem decomposition. However, a significant challenge lies in the modality gap, as LLMs primarily process textual information, whereas robotic systems require multi-modal data (e.g., vision, audio) for effective operation. Multi-modal models such as CLIP \cite{radford2021learning} or BLIP \cite{li2022blip}, highlighted by \cite{wang2024large}, integrate data from various domains with different extent, enriching the robot's perception and ``understanding'' of its environment. 

The choice and combination of modalities are critical for the robustness of robotic systems. According to \cite{lagerstedt2023multiple}, redundant modalities, such as dual cameras, enhance positional awareness and obstacle detection by providing multiple sources of the same type of information, similar to human binocular vision. Additive modalities, like combining depth scanners with cameras, offer complementary information that allows robots to navigate even in challenging conditions like low light. This approach not only improves the robot's environmental perception but also its ability to handle disturbances.

These studies collectively emphasise the importance of multi-modal inputs and robust simulation environments in developing effective and reliable robotic systems. Hence, we will be modifying selected tasks from the ALFRED benchmark in AI2Thor for our study on disturbances. \\

\paragraph{Selected Models}

Our study aims to assess the robustness of different multi-modal models for robotic agents with respect to different modalities. Hence, we briefly summarise the works of those papers that introduce the models which we plan to assess. Selection problems and arguments for choosing these models are outlined in the methodology (Sec. \ref{sec:methods}).

The Hierarchical Language-conditioned
Spatial Model (\textbf{HLSM}) bridges the gap between high-level natural language instructions and robot actions over long execution periods \cite{blukis2022persistent}. By maintaining a continuous spatial-semantic map, HLSM allows robotic agents to perform hierarchical reasoning, decomposing complex tasks into manageable sub-tasks. This approach, which avoids step-by-step instructions, has demonstrated state-of-the-art performance on the ALFRED benchmark, making it highly effective for executing long-term tasks based on abstract language commands.

\textbf{FILM} (Following Instructions in Language with Modular methods) utilises structured representations to achieve good results \cite{min2021film}. Their model constructs a semantic map of the scene and uses a semantic search policy for exploration, avoiding reliance on expert trajectories and low-level instructions. It achieves state-of-the-art performance (24.46\%), and demonstrates further improvement (26.49\%) with the integration of low-level language. This approach highlights the efficacy of explicit spatial memory and semantic search for robust state tracking and task execution.

Complementing the other models, Embodied BERT (\textbf{EmBERT}) addresses the challenge of grounding language instructions in visual observations and actions for language-guided robots \cite{suglia2021embodied}. EmBERT, a transformer-based model, processes high-dimensional, multi-modal inputs across long temporal horizons for task completion. It bridges the gap between object-centric navigation models and the ALFRED benchmark by incorporating object navigation targets in its training. EmBERT achieves competitive performance on ALFRED, being one of the first transformer-based models to manage long-horizon, dense, multi-modal histories, and to utilise object-centric navigation targets effectively.

\section{Methodology}
\label{sec:methods}

\begin{figure*}[t]
    \centering
    \includegraphics[width=\textwidth]{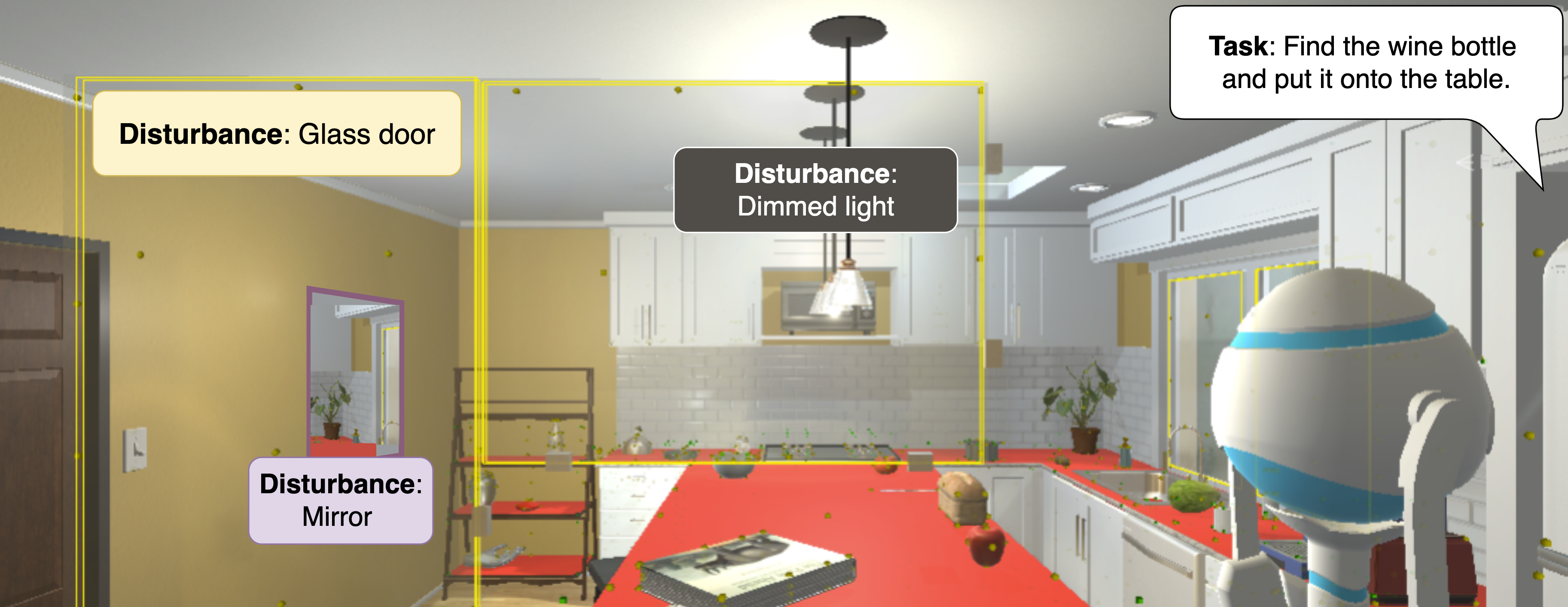}
    \caption{Several state-of-the-art models for assistive robots are evaluated in the AI2Thor environment for their robustness against challenging \textit{disturbances}, such as dimmed lighting or encountering a glass door (see the above image).}
    \label{fig:dists}
\end{figure*}

\subsection{Model Selection}

Selecting models for (simulated) robotic tasks posed several challenges. Few models are open source, and many are outdated, requiring significant time to reproduce original results due to poor documentation and outdated packages. Additionally, top-performing models on the ALFRED leaderboard\footnote{\url{https://leaderboard.allenai.org/alfred/submissions/public}} are often inaccessible. Our objective was to choose models with diverse architectures, modalities, and interfaces with the environment. However, the pool of potential multimodal robotic agents was limited, and migrating models trained on different platforms to a common simulator proved unfeasible. This is because agent implementations vary across platforms and tasks, tailored specifically to the inputs and environments they were originally designed for. The choice for HLSM, FILM and EmBERT was therefore motivated by using open-source, well-performing models on the ALFRED benchmark. 

\subsection{Disturbances}

We define a \textit{disturbance} as any external factor or alteration in the environment that directly limits or disables one or more properties of one or more modalities, thus challenging the model's ability to generalise and respond effectively to new, unseen scenarios. In the following, we list the current set of disturbances along with their explanations (see Fig. \ref{fig:dists}): 

\begin{itemize}
    \item \textbf{Dimmed light}: The same task is executed under different light conditions. From brightly lit to no-light.

    \item \textbf{Glass door}: The room is blocked by a glass door. Blockage of moves can detect the wall, vision may not. 

    \item \textbf{Mirror}: A mirror on a wall will provide redundant ant visual information. A depth sensor may pick up on it. 
    
\end{itemize}

This initial list is not complete and we aim to develop further disturbances targeting different modalities.

\subsection{Tasks Selection}

In order to implement disturbances to the environment, the tasks for the robots need to fulfil two criteria: i) the task must actually be disturbed by our modification with respect to some modality (vision, depth etc.). ii) the model must have at least some good ability to solve the task, because otherwise the impact of the disturbance would likely be immeasurable. 
The tasks used in our study are selected from the seen test set\footnote{\url{https://github.com/askforalfred/alfred/tree/master/data/json_2.1.0/tests_seen}} of the ALFRED Benchmark, focusing on a single floorplan. This approach simplifies the process by allowing changes to one floorplan, making it easier to observe the effects of various disturbances. The selected tasks encompass multiple types classified within the ALFRED Benchmark, such as ``pick two objects and place,'' ``heat and place,'' and more. Specifically, our selection of tasks are described by the following Task IDs:
\begin{itemize}
    \item Task \#1: \texttt{trial\_T20190906\_234933\_757762}
    \item Task \#2: \texttt{trial\_T20190906\_200537\_899818}
    \item Task \#3: \texttt{trial\_T20190907\_163216\_451970}
    \item Task \#4: \texttt{trial\_T20190908\_062227\_162609}
    \item Task \#5: \texttt{trial\_T20190908\_200539\_115276}
    \item Task \#6: \texttt{trial\_T20190908\_062304\_008535}

\end{itemize} 

Each task includes three goal-instruction pairs. For example, in \texttt{trial\_T20190908\_062227\_162609}, the goals are:
\begin{itemize}
    \item ``Put a warm plate in the sink.''
    \item ``Put a heated white plate in the sink.''
    \item ``Warm a plate and put it in the sink.''
\end{itemize}

Each goal is paired with its corresponding instructions. While the model has seen the floorplan during training, it has not been trained on scenes with disturbances, adding an extra layer of challenge. The tasks from the ALFRED Benchmark are selected for their complexity, requiring the virtual robotic agent to demonstrate a range of skills such as navigation, object interaction, memorisation, and the application of world or commonsense knowledge.

\subsection{Evaluation Metric}

The evaluation of task performance is based on two primary metrics: \textbf{i) Task Success}: The percentage of successfully completed tasks.
\textbf{ii) Goal Condition Success}: The percentage of successfully completed sub-goals within each task.

To ensure robustness in the evaluation, each task is run with different starting positions of objects and agent. Initially, there were three runs per task, corresponding to one run per goal-instruction pair. To enhance the evaluation, three additional random initial agent positions are selected, resulting in 12 runs per task. Each task is then also evaluated under disturbance-conditions, resulting in 12 runs per disturbed task as well. This comprehensive approach ensures that the evaluation captures a wide range of possible scenarios and agent behaviours.

\section{Current Progress}

The current state of the research includes the extension of the dataset with three additional starting positions for the robot, which have been randomly generated and tested for necessary minimal distances. For each task, there are now 4 starting positions with the three different goal-instructions for six tasks, which aggregates to a total of 72 episodes. These have been tested with and without a glass wall disturbance within the task scenario two models (FILM and HLSM). 

\subsection{Preliminary Results}

Quantitative results were collected automatically to determine the success rate, while goal states were manually verified by checking intermediate sub-goal success. These results, aggregated across all starting positions and tasks for two models under two conditions (with and without a glass wall disturbance), are presented in the following Table:

\begin{table}[h]
    \centering
    \begin{tabular}{l r r}
        \textbf{Condition} & \textbf{Success Rate} & \textbf{Goal Condition} \\ \hline
        FILM baseline& 16.67\% &39.41\%\\
FILM glass wall disturbance&16.67\%&31.28\%\\
HLSM baseline&15.28\%&44.02\%\\
HLSM glass wall disturbance&8.33\%&26.13\%\\

    \end{tabular}
    \label{tab:res}
\end{table}

The presence of a glass wall disturbance leads to a lower overall Goal Condition Rate (percentage of successful subgoals). The HLSM model appears more affected by the glass wall compared to the FILM model, with 17.89\% of subtasks remaining unfinished due to the disturbance.

This trend does not hold consistently across individual tasks. For one task, the FILM model successfully completed the final subgoal three times, overcoming previous failures. Conversely, the HLSM model detected a microwave for the first time in the glass wall scenario - an otherwise difficult subgoal. For HLSM, the addition of ground truth depth information increased performance in glass wall scenes to a success rate (SR) of 12.50\% and a Goal Condition (GC) of 35.36\%. This aligns with ablation studies by the authors of HLSM and FILM, indicating that ground truth depth data is particularly beneficial in unfamiliar environments, such as when encountering new objects or obstacles in an already familiar room.

\begin{figure}
    \centering
    \includegraphics[width=.35\textwidth]{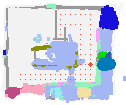}
    \caption{Semantic map produced by FILM during Task \#4, where the agent (red marker) registers the glass wall as an obstacle (grey vertical line at top center) and navigates away from it (red dots) without getting stuck}
    \label{fig:semMap}
\end{figure}

Qualitative observations indicate that the glass wall often causes agents to become stuck or exhibit erratic behaviour. In the ALFRED benchmark, which limits agents to ten failed actions, this threshold is frequently reached sooner in the glass wall condition due to agents repeatedly bumping into the wall. An exception to this can be observed for FILM in Task \#4, where the agent registers the wall and navigates away from it (see Figure \ref{fig:semMap} for the semantic map produced during the task).

\subsection{Next Steps}

Future work will focus on several areas. First, we evaluate scenes containing mirrors, as a new form of visual disturbance, to understand their impact on model performance. Second, introduce additional obstacles and disturbances to assess their effects comprehensively. Third, investigate how exactly the starting positions influence individual results of tasks. Lastly, we plan to consider comparing the effects of disturbances with results from ALFRED tasks conducted on unknown floorplans to provide a broader context for the effects of disturbances.

\section{Conclusion}

The presented project challenges existing multimodal models for robotic tasks by introducing disturbances in virtual environments. Many models under development claim that training on familiar tasks enhances their generalisation capabilities to unfamiliar tasks. These models argue that not every object needs to be seen, nor every room explored, due to the extensive understanding provided by large language and vision models of objects, scenes, and spatial conditions. However, the extent to which different sensory input streams influence task performance remains understudied. While ablation studies offer some insights, our research specifically examines the disturbance of selected or combined inputs. Although our work is still ongoing and results are pending, we believe that the methodology proposed in this study can guide  and encourage researchers to consider disturbances during development.

Many of our disturbances are derived from real-life scenarios, such as mirrors, glass doors, and lights being turned off. Nevertheless, further exploration is needed to understand how environmental factors can disrupt workflows and to introduce more realistic and challenging disturbances. Therefore, we aim to utilise the Geriatronics Summit as an opportunity to discuss, and explore scenarios for assistive robots in greater depth.


%

\ifCLASSOPTIONcaptionsoff
  \newpage
\fi



\bibliographystyle{IEEEtran}
\bibliography{main}

\begin{thebibliography}{10}
\providecommand{\url}[1]{#1}
\csname url@samestyle\endcsname
\providecommand{\newblock}{\relax}
\providecommand{\bibinfo}[2]{#2}
\providecommand{\BIBentrySTDinterwordspacing}{\spaceskip=0pt\relax}
\providecommand{\BIBentryALTinterwordstretchfactor}{4}
\providecommand{\BIBentryALTinterwordspacing}{\spaceskip=\fontdimen2\font plus
\BIBentryALTinterwordstretchfactor\fontdimen3\font minus \fontdimen4\font\relax}
\providecommand{\BIBforeignlanguage}[2]{{%
\expandafter\ifx\csname l@#1\endcsname\relax
\typeout{** WARNING: IEEEtran.bst: No hyphenation pattern has been}%
\typeout{** loaded for the language `#1'. Using the pattern for}%
\typeout{** the default language instead.}%
\else
\language=\csname l@#1\endcsname
\fi
#2}}
\providecommand{\BIBdecl}{\relax}
\BIBdecl

\bibitem{ye2022rcare}
R.~Ye, W.~Xu, H.~Fu, R.~K. Jenamani, V.~Nguyen, C.~Lu, K.~Dimitropoulou, and T.~Bhattacharjee, ``Rcare world: A human-centric simulation world for caregiving robots,'' in \emph{2022 IEEE/RSJ International Conference on Intelligent Robots and Systems (IROS)}.\hskip 1em plus 0.5em minus 0.4em\relax IEEE, 2022, pp. 33--40.

\bibitem{kolve2017ai2}
E.~Kolve, R.~Mottaghi, W.~Han, E.~VanderBilt, L.~Weihs, A.~Herrasti, M.~Deitke, K.~Ehsani, D.~Gordon, Y.~Zhu \emph{et~al.}, ``Ai2-thor: An interactive 3d environment for visual ai,'' \emph{arXiv preprint arXiv:1712.05474}, 2017.

\bibitem{james2020rlbench}
S.~James, Z.~Ma, D.~R. Arrojo, and A.~J. Davison, ``Rlbench: The robot learning benchmark \& learning environment,'' \emph{IEEE Robotics and Automation Letters}, vol.~5, no.~2, pp. 3019--3026, 2020.

\bibitem{bucker2022reshaping}
A.~Bucker, L.~Figueredo, S.~Haddadinl, A.~Kapoor, S.~Ma, and R.~Bonatti, ``Reshaping robot trajectories using natural language commands: A study of multi-modal data alignment using transformers,'' in \emph{2022 IEEE/RSJ International Conference on Intelligent Robots and Systems (IROS)}.\hskip 1em plus 0.5em minus 0.4em\relax IEEE, 2022, pp. 978--984.

\bibitem{bucker2023latte}
A.~Bucker, L.~Figueredo, S.~Haddadin, A.~Kapoor, S.~Ma, S.~Vemprala, and R.~Bonatti, ``Latte: Language trajectory transformer,'' in \emph{2023 IEEE International Conference on Robotics and Automation (ICRA)}.\hskip 1em plus 0.5em minus 0.4em\relax IEEE, 2023, pp. 7287--7294.

\bibitem{zhang2023lohoravens}
S.~Zhang, P.~Wicke, L.~K. {\c{S}}enel, L.~Figueredo, A.~Naceri, S.~Haddadin, B.~Plank, and H.~Sch{\"u}tze, ``Lohoravens: A long-horizon language-conditioned benchmark for robotic tabletop manipulation,'' \emph{arXiv preprint arXiv:2310.12020}, 2023.

\bibitem{wicke2023towards}
P.~Wicke, L.~K. {\c{S}}enel, S.~Zhang, L.~Figueredo, A.~Naceri, S.~Haddadin, and H.~Sch{\"u}tze, ``Towards language-based modulation of assistive robots through multimodal models,'' in \emph{presentations at the Geriatronics Summit 2023: arXiv preprint arXiv:2306.14830}, 2023.

\bibitem{radford2021learning}
A.~Radford, J.~W. Kim, C.~Hallacy, A.~Ramesh, G.~Goh, S.~Agarwal, G.~Sastry, A.~Askell, P.~Mishkin, J.~Clark \emph{et~al.}, ``Learning transferable visual models from natural language supervision,'' in \emph{International conference on machine learning}.\hskip 1em plus 0.5em minus 0.4em\relax PMLR, 2021, pp. 8748--8763.

\bibitem{brohan2023can}
A.~Brohan, Y.~Chebotar, C.~Finn, K.~Hausman, A.~Herzog, D.~Ho, J.~Ibarz, A.~Irpan, E.~Jang, R.~Julian \emph{et~al.}, ``Do as i can, not as i say: Grounding language in robotic affordances,'' in \emph{Conference on robot learning}.\hskip 1em plus 0.5em minus 0.4em\relax PMLR, 2023, pp. 287--318.

\bibitem{driess2023palm}
D.~Driess, F.~Xia, M.~S. Sajjadi, C.~Lynch, A.~Chowdhery, B.~Ichter, A.~Wahid, J.~Tompson, Q.~Vuong, T.~Yu \emph{et~al.}, ``Palm-e: An embodied multimodal language model,'' in \emph{International Conference on Machine Learning}.\hskip 1em plus 0.5em minus 0.4em\relax PMLR, 2023, pp. 8469--8488.

\bibitem{padalkar2023open}
A.~Padalkar, A.~Pooley, A.~Jain, A.~Bewley, A.~Herzog, A.~Irpan, A.~Khazatsky, A.~Rai, A.~Singh, A.~Brohan \emph{et~al.}, ``Open x-embodiment: Robotic learning datasets and rt-x models,'' \emph{arXiv preprint arXiv:2310.08864}, 2023.

\bibitem{shah2023lm}
D.~Shah, B.~Osi{\'n}ski, S.~Levine \emph{et~al.}, ``Lm-nav: Robotic navigation with large pre-trained models of language, vision, and action,'' in \emph{Conference on robot learning}.\hskip 1em plus 0.5em minus 0.4em\relax PMLR, 2023, pp. 492--504.

\bibitem{shridhar2020alfred}
M.~Shridhar, J.~Thomason, D.~Gordon, Y.~Bisk, W.~Han, R.~Mottaghi, L.~Zettlemoyer, and D.~Fox, ``Alfred: A benchmark for interpreting grounded instructions for everyday tasks,'' in \emph{Proceedings of the IEEE/CVF conference on computer vision and pattern recognition}, 2020, pp. 10\,740--10\,749.

\bibitem{chen2023towards}
L.~Chen, Y.~Zhang, S.~Ren, H.~Zhao, Z.~Cai, Y.~Wang, T.~Liu, and B.~Chang, ``Towards end-to-end embodied decision making with multi-modal large language model: Explorations with gpt4-vision and beyond,'' in \emph{NeurIPS 2023 Foundation Models for Decision Making Workshop}, 2023.

\bibitem{wei2022chain}
J.~Wei, X.~Wang, D.~Schuurmans, M.~Bosma, F.~Xia, E.~Chi, Q.~V. Le, D.~Zhou \emph{et~al.}, ``Chain-of-thought prompting elicits reasoning in large language models,'' \emph{Advances in neural information processing systems}, vol.~35, pp. 24\,824--24\,837, 2022.

\bibitem{li2022blip}
J.~Li, D.~Li, C.~Xiong, and S.~Hoi, ``Blip: Bootstrapping language-image pre-training for unified vision-language understanding and generation,'' in \emph{International conference on machine learning}.\hskip 1em plus 0.5em minus 0.4em\relax PMLR, 2022, pp. 12\,888--12\,900.

\bibitem{wang2024large}
J.~Wang, Z.~Wu, Y.~Li, H.~Jiang, P.~Shu, E.~Shi, H.~Hu, C.~Ma, Y.~Liu, X.~Wang \emph{et~al.}, ``Large language models for robotics: Opportunities, challenges, and perspectives,'' \emph{arXiv preprint arXiv:2401.04334}, 2024.

\bibitem{lagerstedt2023multiple}
E.~Lagerstedt and S.~Thill, ``Multiple roles of multimodality among interacting agents,'' \emph{ACM Transactions on Human-Robot Interaction}, vol.~12, no.~2, pp. 1--13, 2023.

\bibitem{blukis2022persistent}
V.~Blukis, C.~Paxton, D.~Fox, A.~Garg, and Y.~Artzi, ``A persistent spatial semantic representation for high-level natural language instruction execution,'' in \emph{Conference on Robot Learning}.\hskip 1em plus 0.5em minus 0.4em\relax PMLR, 2022, pp. 706--717.

\bibitem{min2021film}
S.~Y. Min, D.~S. Chaplot, P.~K. Ravikumar, Y.~Bisk, and R.~Salakhutdinov, ``Film: Following instructions in language with modular methods,'' in \emph{International Conference on Learning Representations}, 2021.

\bibitem{suglia2021embodied}
A.~Suglia, Q.~Gao, J.~Thomason, G.~Thattai, and G.~Sukhatme, ``Embodied bert: A transformer model for embodied, language-guided visual task completion,'' \emph{arXiv preprint arXiv:2108.04927}, 2021.

\end{thebibliography}
%


%




\end{document}